\documentclass[11pt,a4paper]{article}

\usepackage[T1]{fontenc}
\usepackage[utf8]{inputenc}
\usepackage{lmodern}
\usepackage[margin=2.5cm]{geometry}
\usepackage{amsmath,amssymb,bm}
\usepackage{graphicx}
\usepackage{booktabs,tabularx,multirow}
\usepackage[table]{xcolor}
\usepackage{caption}
\usepackage[numbers,sort&compress]{natbib}
\usepackage{microtype}
\usepackage{authblk}
\usepackage{siunitx}
\usepackage{enumitem}
\usepackage[hidelinks]{hyperref}
\usepackage[capitalise,noabbrev]{cleveref}

\graphicspath{{figures/}}
\setlist{itemsep=2pt,topsep=4pt}

\crefname{equation}{Eq.}{Eqs.}
\Crefname{equation}{Equation}{Equations}

\newcommand{\nrmse}{\mathrm{nRMSE}}
\newcommand{\nmae}{\mathrm{nMAE}}

\title{Explainable AI for Solar PV Power Forecasting:\\ From Model Selection to Trustworthy, Reproducible Deployment}

\author[1,3]{Nuzhat Khan}
\author[2]{Muhammad Paend Bakht\thanks{Corresponding author: paend.bakht@mu.ie}}
\author[1]{Ab Al-Hadi Ab Rahman}
\author[3]{Mohamed Khalafalla Hassan}
\author[3]{Paul Stynes}
\author[4]{Muhammad Anique Aslam}

\author[4]{Terence O'Donnell}
\author[5]{Mohammad Riadh}

\affil[1]{Faculty of Electrical Engineering, Universiti Teknologi Malaysia, Malaysia}
\affil[2]{
School of Business, Maynooth University, Ireland}
\affil[3]{School of Computing, National College of Ireland, Ireland}

\affil[4]{School of Electrical and Electronic Engineering, University College Dublin, Ireland}
\affil[5]{School of Mechanical and Manufacturing Engineering, Dublin City University, Ireland}
\date{}

\begin{document}
\maketitle

\begin{abstract}
\noindent
Machine learning (ML) is now a standard tool for forecasting the output of solar photovoltaic (PV) plants, yet published accuracy figures are often hard to compare and harder to trust. Many studies validate with random splits, report no reference forecast, and present feature attributions without saying what they can and cannot show. This chapter reviews the main families of ML methods for PV power forecasting and resource assessment, from tree ensembles to deep and foundation models, and sets out a reproducible workflow that combines automated model search with explainable AI. The workflow is then tested on an openly available dataset: a 16.32\,kWp plant in Porto, Portugal, forecast one day ahead from a 4\,km numerical weather prediction grid. Over a full test year, the best models reach a normalised root mean square error (nRMSE) of 11.6\% of installed capacity, about half the error of smart persistence. The choice of input features changed accuracy far more than the choice among strong algorithms, and shuffled cross-validation was mildly optimistic. After conformal calibration, the nominal 80\% prediction intervals covered 81.7\% of observations, against 64.5\% before. SHAP rankings stayed stable when the model was retrained on different periods but shifted when correlated inputs were removed, so attributions should be read as properties of a model rather than of the plant. The chapter closes with practical lessons and open problems, including deployment on edge hardware and the effect of a changing climate.

\medskip\noindent
\textbf{Keywords:} solar photovoltaic; automated machine learning; explainable AI; SHAP; numerical weather prediction; probabilistic forecasting
\end{abstract}

\section{Introduction}\label{sec:intro}
The world had more than 2.2\,TW of solar photovoltaic (PV) installed by the end of 2024, over 600\,GW of it added in that year alone. In the same year, PV supplied more than 10\% of global electricity demand for the first time \citep{iea2025}. At this scale, PV affects how power systems are operated. Its output follows the sun and the clouds rather than demand, so grid operators, battery schedulers and market participants who bid a day ahead all need a reliable estimate of the next day’s generation \citep{inman2013,antonanzas2016}.

Forecasting PV power is now part of routine grid operation rather than a research exercise. Transmission operators, aggregators and energy communities use forecasts every day, and the quality of those forecasts affects reserve costs, curtailment and the price at which energy is sold \citep{yang2022}. Machine learning (ML) has become one of the most common ways to produce them. Tree ensembles, neural networks and, more recently, transformers and pre-trained foundation models are routinely reported to outperform physical and statistical approaches.

The PV forecasting literature has two weak points. The first is trust. Operators are expected to act on forecasts produced by models they cannot inspect, and this has prompted calls either to make black-box models explainable or to replace them with models that are interpretable by design \citep{rudin2019,machlev2022}. Explainable AI (XAI) methods such as SHapley Additive exPlanations (SHAP) are now common in energy studies, yet their limitations are rarely discussed. The second weak point is reproducibility. Models are often validated with random train--test splits that leak information across time, compared only with other machine learning models rather than with a simple reference forecast, and trained on private data that others cannot access \citep{kapoor2023,yang2020verif}. Reported errors then appear smaller than they really are, and results from different studies cannot be compared on equal terms.

A previous study combined automated machine learning (AutoML) with SHAP to predict the daily output of a rooftop plant in tropical Malaysia from five weather variables \citep{bakht2025}. Extra Trees was the most accurate model, and SHAP ranked irradiance and temperature as the most influential inputs. The study, however, used same-day weather observations, random validation and no reference forecast. This research tests the same workflow under stricter conditions: a temperate climate, a true day-ahead forecast driven by numerical weather predictions (NWP), time-ordered validation, reference baselines, calibrated uncertainty and a closer examination of what the explanations can support.

The chapter first reviews the main AI methods for PV power forecasting and resource assessment, noting where each performs well and where it falls short. It then describes a reproducible workflow that links automated model search with explainable AI and identifies the design decisions that most affect the result. Finally, it applies this workflow to an open dataset \citep{andrade2017,andrade2020data}, so that every number reported in the case study can be checked independently.

\Cref{sec:problem} defines the forecasting problem and its data sources. \Cref{sec:methods} reviews forecasting methods, and \cref{sec:xai} covers the tools used to explain them. The workflow is presented in \cref{sec:workflow} and applied to the case study in \cref{sec:case}. \Cref{sec:challenges} discusses deployment challenges and open problems, and \cref{sec:conclusion} concludes.
\section{The PV forecasting problem}\label{sec:problem}

A forecast is useful only if it fits the decision it supports. Who uses it, and when, determines the horizon, the time step and the form of the output \citep{antonanzas2016,yang2022}. \Cref{tab:horizons} summarises the settings that appear most often in practice.

\begin{table}[htbp]
\centering
\caption{Forecast horizons for PV power and their main uses.}\label{tab:horizons}
\small
\begin{tabularx}{\textwidth}{@{}l l X X@{}}
\toprule
Horizon & Resolution & Main inputs & Main uses \\
\midrule
Intra-hour (s to 1\,h) & 1\,s--15\,min & Sky cameras, on-site sensors, recent output & Ramp control, storage dispatch, inverter set-points \\
Intra-day (1--6\,h) & 15\,min--1\,h & Satellite cloud motion, recent output, NWP & Intra-day trading, reserve scheduling \\
Day-ahead (24--48\,h) & 15\,min--1\,h & NWP & Day-ahead market bids, unit commitment, community energy planning \\
Medium term (days--weeks) & 1\,h--1\,day & NWP ensembles, climatology & Maintenance planning, hydro and fuel scheduling \\
Long term (months--years) & 1\,day--1\,year & Satellite and reanalysis archives, climate data & Resource assessment, plant design, project finance \\
\bottomrule
\end{tabularx}
\end{table}

\subsection{Estimation versus forecasting}

Many studies described as forecasts are, strictly speaking, estimates. They map the weather observed at time $t$ to PV output at the same time $t$. A model of this kind is useful for filling gaps in a monitoring record or for spotting a plant that underperforms, but it cannot be used to plan ahead because the weather at time $t$ is not yet known. A true forecast uses only information available when it is issued, such as an NWP run, a satellite image or past output.

The distinction has a large effect on accuracy. An estimation model fed with measured irradiance can reach very high scores, since irradiance alone explains most of the variation in PV output. A day-ahead model has to rely on forecast irradiance, which carries the errors of the weather model, particularly for cloud \citep{lorenz2009,perez2013}. Scores from the two settings should therefore not be compared. The earlier study \citep{bakht2025} was an estimation study in this sense; the case study in \cref{sec:case} uses a genuine day-ahead set-up.

\subsection{Point and probabilistic forecasts}

A point forecast gives one value per time step. A probabilistic forecast gives a distribution, usually summarised by quantiles or a prediction interval. System operators increasingly ask for the second kind, because the cost of an error is not symmetric and reserves are sized from the range of plausible outcomes \citep{hong2016,pinson2013}. Point forecasts are scored with error measures such as the root mean square error (RMSE) and the mean absolute error (MAE). Probabilistic forecasts need proper scoring rules, for instance the pinball loss or the continuous ranked probability score, together with checks of calibration \citep{gneiting2007}.

\subsection{Data sources}

PV forecasting models draw on four main types of data. First is the plant’s own monitoring record, usually inverter or meter output, often accompanied by module temperature and plane-of-array irradiance. This record is the forecasting target and, for short horizons, also one of the strongest inputs. It should be checked carefully before use, because clipping, curtailment, outages and soiling can distort the measured output \citep{sengupta2021}.
For horizons beyond a few hours, numerical weather prediction becomes the main input. Global and regional models such as the Integrated Forecasting System of the European Centre for Medium-Range Weather Forecasts (ECMWF IFS), the Global Forecast System (GFS) and the Weather Research and Forecasting (WRF) model \citep{skamarock2019} provide forecasts of irradiance, cloud cover and temperature from several hours to several days ahead. Beyond roughly six hours ahead, most forecast skill comes from these models, even though their irradiance forecasts carry systematic biases that machine learning can partly correct \citep{lorenz2009,perez2013}.
Satellite and reanalysis products fill a different role. Geostationary satellites capture cloud fields every 5 to 15 minutes, while reanalysis datasets such as ERA5 and NASA POWER offer long records without gaps. These sources suit intra-day forecasting and resource assessment best. For the shortest horizons, all-sky cameras installed near the plant track approaching clouds and support forecasts only minutes ahead \citep{sun2019}.
\subsection{Resource assessment}
Resource assessment answers a longer-term question than operational forecasting. Rather than estimating tomorrow’s power, it estimates how much energy a plant will produce over its lifetime and how much that total may vary from year to year. Such estimates rely on long satellite or reanalysis records, typical meteorological years and exceedance values (P50, P90), which lenders use to decide whether to finance a project \citep{sengupta2021}. The same estimates are used to size off-grid and hybrid systems, for example PV-based hybrid systems designed to maintain supply during load shedding \citep{bakht2022plos,bakht2024access}. Here, AI is mainly used to correct satellite estimates with ground measurements, fill gaps in records and downscale climate projections. Many of the methods overlap with operational forecasting, but they are judged differently: what matters is bias over many years rather than error in individual hours. The remainder of this chapter focuses on operational forecasting, which is the setting of the case study, and returns to resource assessment in \cref{sec:challenges}.

\section{Critical review of AI methods}\label{sec:methods}

No single method wins on every PV dataset. Reviews of the field agree that the choice of inputs, horizon and validation scheme often matters more than the algorithm, and that reported gains shrink when methods are compared on the same data under the same rules \citep{antonanzas2016,yang2022,voyant2017}. The main method families are reviewed below and summarised in \cref{tab:methods}, together with their strengths and limitations.

\subsection{Reference forecasts and skill}\label{sec:reference}

Any forecast should be compared with a simple reference, and the skill score is the usual way to express that comparison:
\begin{equation}
  S = 1 - \frac{\mathrm{RMSE}_{\mathrm{model}}}{\mathrm{RMSE}_{\mathrm{ref}}},
  \label{eq:skill}
\end{equation}
where $\mathrm{RMSE}_{\mathrm{model}}$ is the root mean square error of the forecast being assessed and $\mathrm{RMSE}_{\mathrm{ref}}$ that of the reference over the same period. $S=0$ means no improvement on the reference, $S=1$ a perfect forecast, and negative values a forecast worse than the reference. Because the reference absorbs much of the difference between sites and climates, skill scores are easier to compare across studies than raw errors, and recent guidance recommends reporting them routinely \citep{yang2020verif}.

The simplest reference is persistence, which assumes that output at the forecast time equals output at the same time on the previous day (or one step earlier for very short horizons). Smart persistence improves on this by persisting the clear-sky index rather than the power itself \citep{pedro2012,yang2020verif}. For a plant of installed capacity $C$, the clear-sky power and the clear-sky index are given as \begin{equation}
  P_{\mathrm{cs}}(t) = \mathrm{PR}\cdot C\,\frac{G_{\mathrm{cs}}(t)}{G_{\mathrm{STC}}},
  \qquad
  k_{\mathrm{c}}(t) = \frac{P(t)}{P_{\mathrm{cs}}(t)},
  \label{eq:kc}
\end{equation}
where $G_{\mathrm{cs}}(t)$ is the clear-sky global horizontal irradiance from a model such as Ineichen--Perez \citep{ineichen2002}, $G_{\mathrm{STC}}=1000$\,W\,m$^{-2}$ is the irradiance at standard test conditions, $\mathrm{PR}$ is a performance ratio that lumps together losses in the modules, wiring and inverter, and $P(t)$ is the measured power. A day-ahead smart persistence forecast for time $t$ then carries over the index observed $h=48$ hours earlier:
\begin{equation}
  \hat{P}_{\mathrm{SP}}(t) = k_{\mathrm{c}}(t-h)\,P_{\mathrm{cs}}(t).
  \label{eq:sp}
\end{equation}
In this way the reference follows the daily and seasonal course of the sun while assuming that cloudiness does not change. Climatology, the average output for the same month and hour, is another common reference. A physical model chain, which converts NWP irradiance into power through models of the sky, the modules and the inverter, is a stronger one \citep{mayer2021}.

\subsection{Classical machine learning}

Support vector regression (SVR), $k$-nearest neighbours (k-NN) and shallow neural networks were among the first ML methods applied widely to PV forecasting \citep{inman2013,voyant2017}. They capture non-linear relations between weather and output and work well on small tabular datasets. SVR training time grows quickly with the number of samples, however, and all three methods are sensitive to input scaling and to their hyper-parameters.

\subsection{Tree ensembles}

Tree ensembles are the workhorse of tabular PV forecasting. A random forest averages many decision trees, each grown on a bootstrap sample of the data \citep{breiman2001}. Extra Trees go a step further and also draw the split thresholds at random, which often lowers variance at a smaller computational cost \citep{geurts2006}. Gradient boosting instead grows trees in sequence, each one fitted to the errors left by the previous ones \citep{friedman2001}. XGBoost and LightGBM are fast, regularised implementations of this idea and have done well in forecasting competitions \citep{chen2016,ke2017,hong2016}.

Their appeal is practical. They need little preprocessing, cope with mixed and correlated inputs, train in seconds to minutes, and admit exact SHAP explanations through the TreeSHAP algorithm \citep{lundberg2020}. Their main weakness is that they cannot extrapolate beyond the range of the training data, which becomes a problem when a plant is extended, when it degrades, or when it meets weather it has not seen before. Random forests and Extra Trees also produce large model files explained in \cref{sec:footprint}.

\subsection{Deep learning}

Deep learning brings several architectures to PV forecasting. Recurrent networks, especially the long short-term memory (LSTM) network, learn temporal dependence directly from sequences of past data \citep{hochreiter1997}. Convolutional networks are better suited to spatial inputs and extract useful features from sky images and satellite cloud fields \citep{sun2019}. Transformers, which rely on the attention mechanism \citep{vaswani2017}, have also been adapted to time series. One example is the Temporal Fusion Transformer, which combines static information, inputs known in advance such as NWP, and past observations, and its attention weights give some insight into what the model uses \citep{lim2021}. Where many PV plants are located close together, graph neural networks can forecast all sites at once and share information between neighbouring plants \citep{simeunovic2022}.

Deep models perform best with large datasets, high-resolution inputs such as images, and many sites at once. For a single site with hourly tabular data their advantage is much less clear. On standard long-horizon benchmarks a simple linear model matched or beat several transformer designs \citep{zeng2023}, and later architectures such as PatchTST regained the lead only after careful changes to the way the series is presented to the network \citep{nie2023}. Deep models also need more data, tuning and computing power, and their explanations are harder to verify.

\subsection{Hybrid and physics-informed models}

Hybrid models combine a physical model chain with ML. In one common design the physical chain produces a first estimate and an ML model corrects its residuals; in another, physical quantities such as clear-sky irradiance and solar geometry are given to the ML model as inputs \citep{mayer2021,antonanzas2016}. Physics-informed neural networks go further and add physical constraints to the training loss \citep{raissi2019,karniadakis2021}. These designs usually extrapolate better and need less data, at the price of more domain work. The case study in \cref{sec:case} uses the lighter of the two approaches: physically derived features inside an ML model.

\subsection{Automated machine learning}

AutoML automates the choice of model, the tuning of its hyper-parameters and, in some tools, part of the feature processing \citep{hutter2019}. Libraries such as PyCaret, auto-sklearn and FLAML \citep{wang2021flaml} search across many model families within a fixed time budget. This saves expert time, reduces personal bias in choosing a model, and puts many candidates through the same comparison. That comparison is valuable because no single learner works best on every problem, a point formalised in the ``no free lunch’’ theorems first stated for optimisation \citep{wolpert1997}. 

AutoML also hides decisions that matter. Most libraries default to random cross-validation, which leaks information in time series unless the user changes it. The time budget limits what is explored, and the result depends on it. And the feature set, which often matters more than the model, is usually left to the user. AutoML is therefore best treated as a search tool inside a careful workflow rather than as a substitute for one; \cref{sec:screening} shows why.

\subsection{Foundation models for time series}

Large models pre-trained on many time series, such as Chronos \citep{ansari2024} and TimesFM \citep{das2024}, can forecast a new series without any training on it (``zero-shot''). This is attractive for new sites with little history. For PV they have two limitations at present: most of them are univariate or make only weak use of external inputs such as NWP, and day-ahead PV skill comes mainly from the weather forecast rather than from the history of the output. Their value for operational PV forecasting is still being established.

\begin{table}[htbp]
\centering
\caption{Main AI method families for PV power forecasting.}\label{tab:methods}
\footnotesize
\renewcommand{\arraystretch}{1.15}
\begin{tabularx}{\textwidth}{@{}>{\raggedright}p{2.5cm} >{\raggedright}p{1.75cm} >{\raggedright}p{2.15cm} >{\raggedright}p{2.15cm} >{\raggedright}p{1.7cm} >{\raggedright\arraybackslash}X@{}}
\toprule
Method family & Data needs & Best suited horizons & Interpretability & Compute cost & Main limitations \\
\midrule
Reference forecasts (persistence, climatology, physical chain) & Very low & All, as baselines & Full & Negligible & Weak beyond a few hours; physical chain needs plant data \\
SVR, $k$-NN, shallow ANN & Low to moderate & Intra-day to day-ahead & Low (post hoc only) & Low to moderate & Sensitive to scaling and tuning; SVR slow on large data \\
Tree ensembles (RF, Extra Trees, XGBoost, LightGBM) & Moderate & Intra-day to day-ahead & Good with TreeSHAP & Low & No extrapolation; large model files (\cref{sec:footprint}) \\
Deep learning (LSTM, CNN, Transformer, GNN) & High & Intra-hour (images) to day-ahead; multi-site & Low to moderate & High & Need large datasets; small gains for single-site data \\
Hybrid and physics-informed & Moderate, plus domain knowledge & All & Moderate to good & Moderate & Need plant metadata and expert set-up \\
AutoML & As for the chosen model & As for the chosen model & As for the chosen model & Set by the search budget & Random splits leak in time series; budget-dependent \\
Time-series foundation models & Very low (zero-shot) & Intra-day; sites with little data & Low & High at inference & Mostly univariate; unproven for day-ahead PV \\
\bottomrule
\end{tabularx}
\end{table}
\section{Explainability for energy forecasting}\label{sec:xai}

People in the energy sector people need different things in an explanation. For instance,  a control-room operator needs to understand why the forecast for tomorrow is low before committing reserves. A plant owner is more interested in which conditions lead to lost output. Regulators and investors want assurance that a model behaves sensibly before they rely on it. Modellers, in turn, need to detect when a model has picked up an artefact of the data instead of the physical behaviour of the plant \citep{machlev2022,arrieta2020}. Because these needs differ, no single explanation method can meet all of them.

\subsection{Main methods}

SHAP splits each prediction into contributions from the input features, using Shapley values from cooperative game theory \citep{lundberg2017}. For a model $f$ and an input vector $\bm{x}$ with $M$ features, the explanation has the additive form
\begin{equation}
  f(\bm{x}) = \phi_0 + \sum_{j=1}^{M} \phi_j(\bm{x}),
  \label{eq:shap}
\end{equation}
where $\phi_0$ is the expected model output over the background data (in effect the average prediction) and $\phi_j(\bm{x})$ is the contribution of feature $j$ to this particular prediction. Each contribution is a Shapley value,
\begin{equation}
  \phi_j(\bm{x}) = \sum_{S \subseteq \mathcal{F}\setminus\{j\}}
  \frac{|S|!\,(M-|S|-1)!}{M!}\,\bigl[f_{S\cup\{j\}}(\bm{x}) - f_S(\bm{x})\bigr],
  \label{eq:shapley}
\end{equation}
in which $\mathcal{F}$ is the set of all features, $S$ runs over the subsets that exclude feature $j$, and $f_S(\bm{x})$ is the expected model output when only the features in $S$ are known. In words, $\phi_j$ is the average change in the prediction when feature $j$ is added, taken over every order in which the features could be added. Averaging $|\phi_j|$ over many predictions gives a global importance ranking. For tree models the TreeSHAP algorithm computes these values exactly and quickly \citep{lundberg2020}, which partly explains why SHAP is now so widely used in PV studies \citep{kuzlu2020,bakht2025}.

 Local Interpretable Model-agnostic Explanations (LIME) explains a single prediction by perturbing the inputs around it and fitting a simple surrogate model to the result \citep{ribeiro2016}. It can be applied to any model, but its explanations depend on how the perturbations are sampled and may differ between runs. Partial dependence plots take a global view and show how the average prediction changes as one feature is varied while the others keep their observed values \citep{friedman2001}. When inputs are correlated, accumulated local effects (ALE) plots are the better choice, since partial dependence then averages over combinations of inputs that never occur in practice \citep{apley2020}.

Permutation importance takes a different approach and asks how much the error increases when the values of one feature are randomly shuffled across the evaluation set \citep{breiman2001,fisher2019}:

\begin{equation}
  I_j = \mathrm{RMSE}\bigl(f;\,\bm{X}^{(\pi_j)}\bigr) - \mathrm{RMSE}\bigl(f;\,\bm{X}\bigr),
  \label{eq:perm}
\end{equation}
where $\bm{X}$ is the evaluation data and $\bm{X}^{(\pi_j)}$ is the same data with column $j$ randomly permuted. In practice $I_j$ is averaged over several permutations. Because it is expressed in the units of the error, it is easy to read. Its weakness is that shuffling creates unrealistic inputs when features are correlated \citep{hooker2021}. Finally, attention weights in transformer models such as the Temporal Fusion Transformer show which inputs and time steps the model attends to \citep{lim2021}. They are useful for inspection but do not measure the effect of an input on the output in the sense of \cref{eq:shap}.

\subsection{Pitfalls in energy applications}\label{sec:pitfalls}

Four problems recur in PV studies, and each of them shows up in the case study in \cref{sec:case}. Correlation between inputs is a particular concern for weather-based models. Forecast irradiance, clear-sky irradiance, solar elevation and temperature all rise and fall with the sun, so they are strongly correlated with each other. The interventional form of \cref{eq:shapley} used by most software treats features as if they could be varied independently, so credit may be divided among correlated inputs in ways that depend on arbitrary choices made during training \citep{aas2021,molnar2022}. In such cases, removing one input can shift its credit to another without any change in accuracy.

A second problem lies in the gap between the model and the physics. SHAP describes how the model reaches its output, not how the plant itself works. A large SHAP value for temperature shows that the model relies on temperature; it does not show that temperature causes the change in output. Physical conclusions, such as an ``optimal temperature range’’ for a plant, therefore need separate physical evidence.

Explanations can also be unstable. Retraining the model on slightly different data, or choosing another model of similar accuracy, may produce a noticeably different ranking \citep{fisher2019}. If an explanation changes each time the model is retrained, it should not be used to guide decisions.

Finally, an explanation is only as relevant as the forecast it explains. SHAP values from an estimation model that uses measured irradiance reveal little about a day-ahead model that relies on forecast irradiance. Every explanation inherits the set-up of its model, including any leakage in the validation.

\subsection{Recommended practice}

Four habits reduce these risks. Report at least two importance measures, for example SHAP and permutation importance, and check whether they agree. Retrain the model on different time windows and compare the rankings. Remove or group redundant inputs and see whether the conclusions change. And break global importance down by season or sky condition, because what drives error on a cloudy winter day is not what drives it on a clear summer one. Where the stakes are high, an interpretable model may be the better choice \citep{rudin2019}; as \cref{sec:case} shows, a well-built linear model on physically meaningful features comes surprisingly close to far more complex ones.

\section{A reproducible AutoML--XAI workflow}\label{sec:workflow}

\begin{figure}[htbp]
  \centering
  \includegraphics[width=\textwidth]{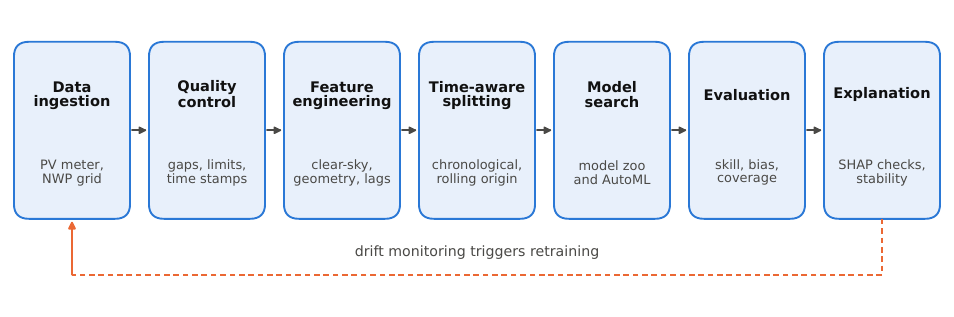}
  \caption{The seven-step AutoML--XAI workflow for PV power forecasting. In operation, drift monitoring feeds back into data ingestion and triggers retraining.}
  \label{fig:workflow}
\end{figure}

The workflow in \cref{fig:workflow} turns the lessons of \cref{sec:problem,sec:methods,sec:xai} into seven steps. It extends the earlier AutoML--SHAP framework \citep{bakht2025} with time-aware validation, reference baselines, uncertainty estimates and checks on the explanations. Each step leaves a record behind it so that another team can repeat the work.

\subsection{Data ingestion and quality control}

The first step gathers plant output, weather inputs and plant metadata (capacity, orientation, commissioning date) and records where each came from and which version was used. Quality control then looks for missing periods, values outside physical limits, frozen sensors, clipping and curtailment \citep{sengupta2021}. Missing targets should be left missing rather than filled, because a filled value rewards the model for reproducing the filling rule. Time stamps need particular care. Plant data may be stamped at the start, middle or end of an interval and recorded in local time, whereas NWP is usually given in UTC and may represent instantaneous or averaged values. A misalignment of one hour can cost as much accuracy as a poor choice of model.

\subsection{Feature engineering}

Physical knowledge enters the model through its features. Clear-sky irradiance from the Ineichen--Perez model \citep{ineichen2002}, computed with an open tool such as pvlib \citep{holmgren2018}, tells the model how much power is possible at each hour. Solar elevation and azimuth describe the geometry. The NWP clear-sky index, that is, forecast irradiance divided by clear-sky irradiance, separates the effect of cloud from the daily and seasonal cycle. Features drawn from neighbouring NWP grid points help when the weather model places a cloud field slightly in the wrong position \citep{andrade2017}. Lagged output is useful only if it is available when the forecast is issued; for a day-ahead forecast made early in the morning, the most recent complete day is usually the day before.

\subsection{Time-aware splitting}

Random $K$-fold cross-validation assumes that the observations are independent, which is not true of time series. Neighbouring hours share the same weather, so a random split puts near-copies of the test hours into the training set and can make the model look better than it is \citep{bergmeir2012,kapoor2023}. Three alternatives avoid this. A chronological hold-out reserves the final period for testing, ideally a full year so that every season is included. Blocked cross-validation keeps whole days or weeks together within one fold. Rolling-origin evaluation retrains the model at regular intervals on all the data available up to that point and tests it on the following period, much as it would be used in operation \citep{tashman2000,hyndman2021}. The same rule applies to hyper-parameter tuning: the search should use time-ordered folds within the training period, and the test period should be used only once.

\subsection{Automated model search}

At this stage, a wide range of models is compared under a single protocol, including linear models, $k$-NN, SVR, single decision trees, random forests, boosted trees and neural networks. An AutoML library can carry out the search, as long as its validation is set to a time-ordered split and its time budget is reported. Running a transparent ``model zoo’’ alongside the AutoML library also reveals how strongly the library’s final choice depends on that budget.

\subsection{Evaluation}

For each hour $t$, let $P_t$ denote the measured power and $\hat{P}_t$ the corresponding forecast, and let $N$ be the number of daytime hours in the evaluation period. Point forecasts are scored with the normalised root mean square error (nRMSE) and normalised mean absolute error (nMAE), both expressed as a percentage of the installed capacity $C$, and with the mean bias error (MBE):
\begin{align}
  \nrmse &= \frac{100}{C}\sqrt{\frac{1}{N}\sum_{t=1}^{N}\bigl(\hat{P}_t - P_t\bigr)^2}, \label{eq:nrmse}\\
  \nmae  &= \frac{100}{C}\,\frac{1}{N}\sum_{t=1}^{N}\bigl|\hat{P}_t - P_t\bigr|, \label{eq:nmae}\\
  \mathrm{MBE} &= \frac{1}{N}\sum_{t=1}^{N}\bigl(\hat{P}_t - P_t\bigr). \label{eq:mbe}
\end{align}
Dividing by installed capacity rather than by mean output makes the errors comparable across plants of different sizes and prevents them from appearing inflated in winter, when mean output is low. A positive MBE indicates that the model over-forecasts on average. The skill score in \cref{eq:skill} relative to smart persistence is also reported
\citep{yang2020verif}. Errors are also broken down by month and by sky condition, since a single yearly average can conceal poor performance on cloudy days.

A quantile forecast $\hat{q}_{\tau,t}$ for probability level $\tau$ (for example $\tau=0.1$ for the 10th percentile) is scored with the pinball loss,
\begin{equation}
  L_\tau = \frac{1}{N}\sum_{t=1}^{N}\max\bigl\{\tau\,(P_t-\hat{q}_{\tau,t}),\;(\tau-1)\,(P_t-\hat{q}_{\tau,t})\bigr\},
  \label{eq:pinball}
\end{equation}
which penalises observations above the quantile with weight $\tau$ and those below it with weight $1-\tau$. For a central interval $[\hat{L}_t,\hat{U}_t]$ with nominal coverage $1-\alpha$, the prediction interval coverage probability (PICP) is the share of observations that fall inside it:
\begin{equation}
  \mathrm{PICP} = \frac{1}{N}\sum_{t=1}^{N}\mathbf{1}\bigl\{\hat{L}_t \le P_t \le \hat{U}_t\bigr\},
  \label{eq:picp}
\end{equation}
where $\mathbf{1}\{\cdot\}$ is the indicator function. It returns one if the measured power $P_t$ falls between the lower bound $\hat{L}_t$ and the upper bound $\hat{U}_t$, and zero if it falls outside them. The PICP is therefore the fraction of hours in which the observation lies inside the interval. A well-calibrated 80\% interval ($\alpha = 0.2$) should give a PICP close to 0.8.

Prediction intervals from raw quantile models are often too narrow, so they contain fewer observations than their nominal coverage suggests. Split-conformal calibration corrects this by widening the intervals by a margin learned on a calibration period that is not used for training \citep{romano2019,vovk2005}. For each of the $n$ calibration hours, a conformity score is computed, and from these scores a margin $\hat{c}$ is obtained:

\begin{equation}
  E_i = \max\bigl\{\hat{L}_i - P_i,\; P_i - \hat{U}_i\bigr\}, \quad
  \hat{c} = \text{the } \bigl\lceil (n+1)(1-\alpha) \bigr\rceil\text{-th smallest value of } \{E_i\}_{i=1}^{n}.
  \label{eq:conformal}
\end{equation}
$E_i$ is positive when the observation falls outside the raw interval and negative when it falls inside. The calibrated interval for a new hour is then
\begin{equation}
  \bigl[\hat{L}_t - \hat{c},\; \hat{U}_t + \hat{c}\bigr],
  \label{eq:cqr}
\end{equation}
clipped to the physical range $[0, C]$. Its coverage is guaranteed when the calibration and test data are exchangeable, an assumption that time series meet only approximately.

\subsection{Explanation}

In the explanation step, the checks recommended in \cref{sec:pitfalls} are put into practice. SHAP values and permutation importance are computed side by side, the stability of the rankings is tested across several retraining windows, and correlated inputs are removed to see whether the conclusions change. Importance is also broken down by season and by sky condition.

\subsection{Recording and monitoring}

For each run, the data versions, code, software versions, random seed and complete results are recorded, following the FAIR (findable, accessible, interoperable, reusable) principles \citep{wilkinson2016}. Once a model is in operation, its inputs and errors are monitored for drift, which may result from soiling, an inverter replacement or an update to the NWP model. When drift is detected, the model is retrained \citep{gama2014}. This feedback loop is shown in \cref{fig:workflow}.
\section{Case study: day-ahead forecasting for a PV plant in Porto}\label{sec:case}

In this section, the workflow described in \cref{sec:workflow} is applied to an openly available dataset. The purpose is to show, with results that others can reproduce, how strongly each design decision affects forecast accuracy, rather than to achieve the best possible score. Details of how to obtain the code and full results are given in \cref{sec:conclusion}.

\subsection{Data and set-up}

The plant is a 16.32\,kWp PV system at the INESC TEC Smart Grid and Electric Vehicle Laboratory in Porto, Portugal (\ang{41.18}\,N, \ang{8.60}\,W) \citep{andrade2017,andrade2020data}. Porto has a temperate oceanic climate with cloudy, wet winters and dry summers, very different from the hot and humid tropical site of the earlier study \citep{bakht2025}. The weather inputs are forecasts from the MeteoGalicia WRF model \citep{skamarock2019} at 4\,km resolution, on a $13\times13$ grid of 169 points centred on the plant. \Cref{tab:setup} summarises the set-up.

\begin{table}[htbp]
\centering
\caption{Case study data and set-up.}\label{tab:setup}
\small
\begin{tabularx}{\textwidth}{@{}l X@{}}
\toprule
Item & Value \\
\midrule
Target & Measured hourly PV power, 28 April 2013 -- 28 June 2016 (\num{26701} hours) \\
NWP variables & Shortwave flux; 2\,m air temperature; total, low, mid and high cloud fraction \\
Forecast task & Day-ahead: the 00\,UTC run on day $D$ forecasts every hour of day $D+1$ (lead times 25--48\,h) \\
Training period & May 2013 -- June 2015 (\num{8321} daytime hours) \\
Test period & July 2015 -- June 2016, one full year (\num{3609} daytime hours) \\
Evaluation & Daytime hours only (solar elevation above \ang{0}); \cref{eq:nrmse,eq:nmae,eq:mbe}, $R^2$ and skill against smart persistence (\cref{eq:skill,eq:sp}) \\
\bottomrule
\end{tabularx}
\end{table}

Quality control found \num{1069} missing hours (3.8\% of the record), most of them in five outages lasting between 3 and 12 days. These hours were left missing. A small number of NWP cloud fractions lay outside the physical range of 0 to 1 (267 values of total cloud fraction, the largest 31.7) and were clipped to that range. One alignment problem turned out to matter. The NWP shortwave flux behaves like an instantaneous value at each time stamp, while the measured power is an average over the hour that ends at the stamp. Averaging the two stamps that bound each hour raised the correlation between forecast irradiance and measured power from 0.889 to 0.912.

Two feature sets were compared. The \emph{weather-only} set mirrors the earlier study and contains the six NWP variables at the plant. The \emph{full} set has 21 features. It adds the hour-mean irradiance; clear-sky irradiance from the Ineichen--Perez model \citep{ineichen2002,holmgren2018}; the NWP clear-sky index; solar elevation and azimuth; the hour of day; and the day of year as a sine and cosine pair. It also includes spatial features (the mean over the $3\times3$ neighbourhood and the standard deviation over the whole grid, for irradiance and for cloud fraction) and lagged features from day $D-1$, the last complete day before the forecast is issued. Rows whose lags were undefined after an outage were dropped.

\subsection{Validation protocol}

\begin{figure}[htbp]
  \centering
  \includegraphics[width=\textwidth]{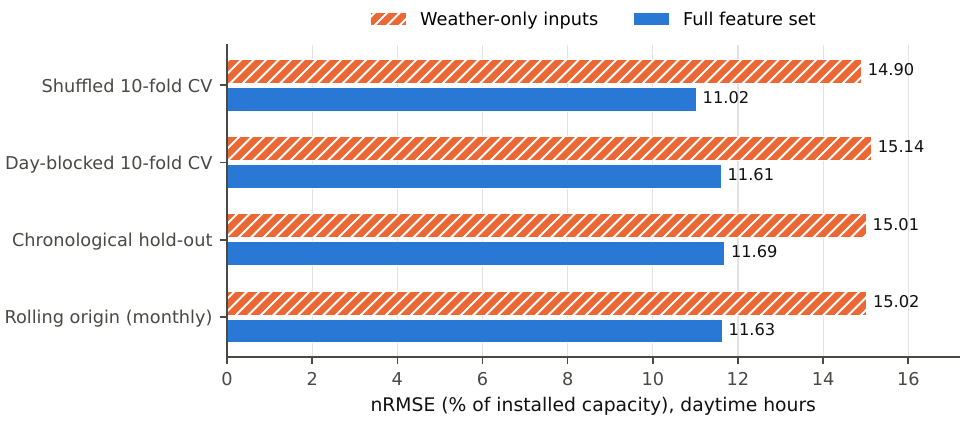}
  \caption{Effect of the validation protocol on the nRMSE of an Extra Trees model, for the weather-only and full feature sets.}
  \label{fig:validation}
\end{figure}

\Cref{fig:validation} compares four ways of validating the same Extra Trees model. With the full feature set, shuffled 10-fold cross-validation gave an nRMSE of 11.02\%. The three protocols that respect time gave 11.61\% (cross-validation with whole days kept together), 11.69\% (chronological hold-out) and 11.63\% (rolling origin with monthly retraining). With weather-only inputs, all four protocols agreed to within 0.25 percentage points.

The optimism of shuffled validation is real but modest here, about 6\% in relative terms. It appears mainly with the full feature set, whose lagged and daily features are shared by neighbouring hours, and in both feature sets it favoured the model. Set-ups whose inputs share more information between hours, such as measured weather or recent output, may show larger gaps. Since the three time-respecting protocols agree closely, a single chronological test year is a reasonable and inexpensive choice for a dataset of this size.

\subsection{Model screening, baselines and skill}\label{sec:screening}

\begin{figure}[htbp]
  \centering
  \includegraphics[width=\textwidth]{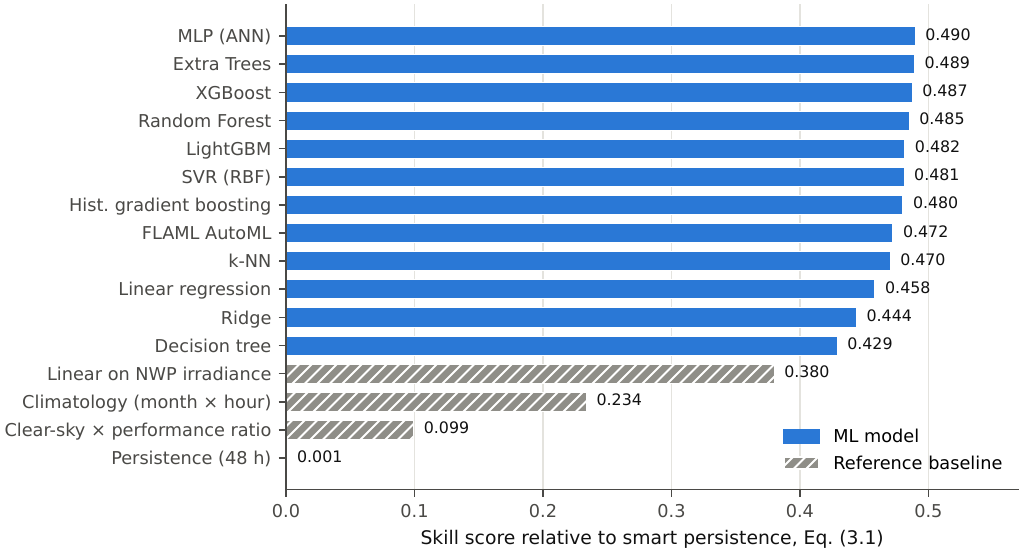}
  \caption{Skill scores of the ML models and reference baselines relative to smart persistence.}
  \label{fig:skill}
\end{figure}

Eleven models by random search were turned with four time-ordered folds inside the training period, and FLAML \citep{wang2021flaml} was given a 7-minute budget.
\Cref{tab:results} and \cref{fig:skill} report the results on the test year.

\begin{table}[htbp]
\centering
\caption{Day-ahead accuracy on the test year (daytime hours, full feature set). Skill is relative to 48-hour smart persistence (\cref{eq:skill,eq:sp}).}\label{tab:results}
\small
\begin{tabular}{@{}l S[table-format=2.2] S[table-format=2.2] S[table-format=1.3] S[table-format=1.3]@{}}
\toprule
Model & {nRMSE (\%)} & {nMAE (\%)} & {$R^2$} & {Skill} \\
\midrule
MLP (two hidden layers, 128 and 64 units) & 11.61 & 8.22 & 0.786 & 0.490 \\
Extra Trees & 11.62 & 8.18 & 0.785 & 0.489 \\
XGBoost & 11.66 & 8.19 & 0.784 & 0.487 \\
Random Forest & 11.71 & 8.16 & 0.782 & 0.485 \\
LightGBM & 11.79 & 8.29 & 0.779 & 0.482 \\
SVR (RBF kernel) & 11.80 & 7.54 & 0.779 & 0.481 \\
Histogram gradient boosting & 11.82 & 8.39 & 0.778 & 0.480 \\
FLAML AutoML (selected Random Forest) & 12.00 & 8.39 & 0.771 & 0.472 \\
$k$-nearest neighbours & 12.05 & 8.58 & 0.769 & 0.470 \\
Linear regression & 12.32 & 9.16 & 0.759 & 0.458 \\
Ridge regression & 12.65 & 9.65 & 0.746 & 0.444 \\
Decision tree & 12.99 & 9.09 & 0.732 & 0.429 \\
\midrule
\multicolumn{5}{@{}l}{\emph{Reference baselines}}\\
Linear regression on NWP irradiance only & 14.10 & 10.74 & 0.684 & 0.380 \\
Climatology (month $\times$ hour) & 17.42 & 13.38 & 0.517 & 0.234 \\
Clear-sky power, \cref{eq:kc} ($\mathrm{PR}=0.854$) & 20.48 & 15.04 & 0.333 & 0.099 \\
Persistence (48\,h) & 22.72 & 15.48 & 0.179 & 0.001 \\
Smart persistence (48\,h), \cref{eq:sp} & 22.74 & 15.49 & 0.178 & 0.000 \\
\bottomrule
\end{tabular}
\end{table}

Four findings stand out. The first is that feature design mattered more than the algorithm. The seven best models are within 0.21 percentage points of each other (11.61--11.82\%). With weather-only inputs, the four tree models from the earlier study scored 14.42\% (Extra Trees), 14.46\% (XGBoost), 14.63\% (Random Forest) and 14.93\% (decision tree). The full feature set therefore improved accuracy by about 2.8 points, more than ten times the spread among the best models.

The second is that Extra Trees remains a strong choice but not a unique one. As in the earlier study \citep{bakht2025}, Extra Trees ranked near the top. This time, however, a tuned MLP and XGBoost matched it, and SVR had the lowest nMAE (7.54\%). The gap between tree models and neural networks that was observed on the Malaysian data did not carry over to this site and set-up, which fits the ``no single winner'' view of \cref{sec:methods}.

The third is that most of the skill comes from the weather model. A linear regression on NWP irradiance alone reached 14.1\%, a skill of 0.38, and the ML models add about 2.5 points on top of that. Persistence, so effective a few minutes ahead, has almost no skill one day ahead.

The fourth is that AutoML with a tight budget did not beat a tuned model zoo. Given 7 minutes on two CPU cores, FLAML settled on a random forest of only seven trees and scored 12.00\%, behind seven of the hand-tuned models. AutoML saves effort, but its answer depends on the budget, and the budget should always be reported.

\subsection{Errors by sky condition and season}

\begin{figure}[htbp]
  \centering
  \includegraphics[width=0.82\textwidth]{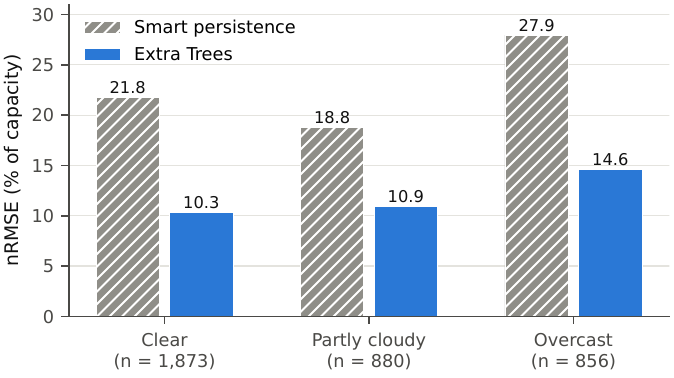}
  \caption{nRMSE of Extra Trees and smart persistence by sky condition on the test year.}
  \label{fig:sky}
\end{figure}

Each test hour was assigned to a sky class using its measured clear-sky index from \cref{eq:kc}: clear ($k_{\mathrm{c}}>0.75$), partly cloudy ($0.35\le k_{\mathrm{c}}\le0.75$) or overcast ($k_{\mathrm{c}}<0.35$). \Cref{fig:sky} shows that Extra Trees reduced the error of smart persistence by 42 to 53\% in every class, but its own error grew from 10.3\% on clear hours to 14.6\% on overcast ones. On overcast hours the model over-forecast by 1.65\,kW on average. When the NWP misses a cloud deck, no amount of post-processing can recover it. On clear hours the model under-forecast by 1.04\,kW. Part of both biases is a selection effect: classifying hours by their measured output picks out exactly those hours that turned out sunnier or cloudier than forecast.

Month by month, the nRMSE of Extra Trees ranged from 9.1\% (June 2016) to 13.0\% (May 2016), and it beat smart persistence in every month of the test year.

\subsection{Probabilistic forecasts}

LightGBM quantile regression was used to produce the 10th, 50th and 90th percentiles. Without calibration the nominal 80\% interval contained only 64.5\% of the test observations (\cref{eq:picp}), with a mean width of 3.45\,kW; the intervals were too narrow. Split-conformal calibration (\cref{eq:conformal,eq:cqr}) on the last six months of the training period, January to June 2015, gave a margin of $\hat{c}=0.32$\,kW. Adding it to each side of the interval raised the coverage to 81.7\% and the mean width to 4.07\,kW. The pinball losses (\cref{eq:pinball}) of the raw quantiles were 0.32, 0.64 and 0.25\,kW for the 10th, 50th and 90th percentiles, and the median forecast had an nRMSE of 11.89\%.

\begin{figure}[htbp]
  \centering
  \includegraphics[width=\textwidth]{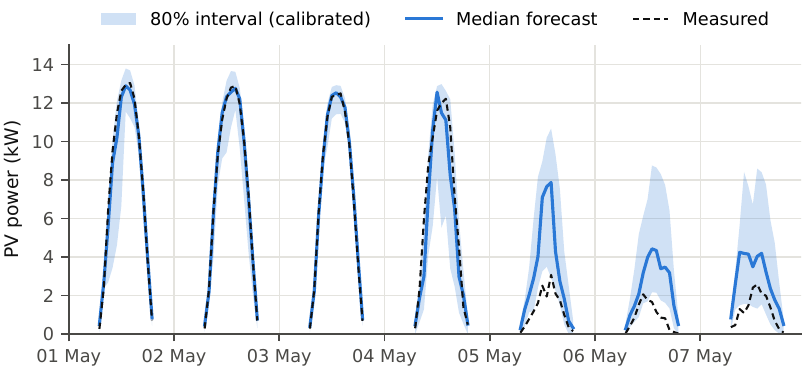}
  \caption{Calibrated day-ahead forecasts for 1--7 May 2016: median and 80\% prediction interval against measured output.}
  \label{fig:prob}
\end{figure}

\Cref{fig:prob} covers a week in which clear days give way to overcast ones. As expected, the interval stays narrow on the clear days and becomes wider on the cloudy ones. On the cloudiest days, however, the median forecast lies well above the measured output, in line with the overcast bias noted earlier.

\subsection{Explanations}

\begin{figure}[htbp]
  \centering
  \includegraphics[width=\textwidth]{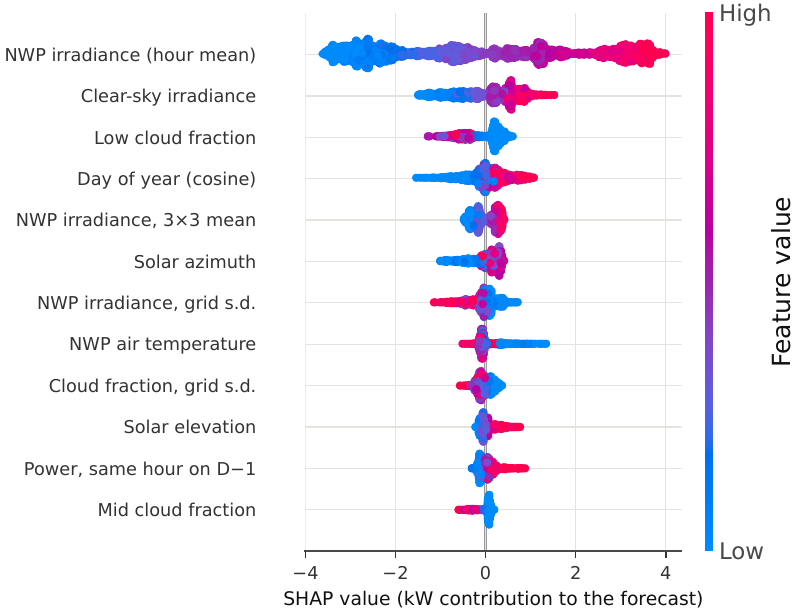}
  \caption{SHAP summary for the LightGBM day-ahead model on the test year. Each dot is one hour; colour shows the value of the feature.}
  \label{fig:shapsummary}
\end{figure}

SHAP values were calculated with TreeSHAP \citep{lundberg2020} for the LightGBM model across all \num{3609} test hours. LightGBM was chosen because its accuracy is close to that of the best models and its explanations can be computed quickly. The overall pattern is shown in \cref{fig:shapsummary}. Hour-mean NWP irradiance is by far the most influential input, with a mean $|\phi_j|$ of 2.10,kW. It is followed by clear-sky irradiance (0.55,kW), low cloud fraction (0.34,kW) and the seasonal term (0.28,kW). Permutation importance (\cref{eq:perm}) ranked the same three features first, in the same order (\cref{tab:importance}).

\begin{figure}[htbp]
  \centering
  \includegraphics[width=\textwidth]{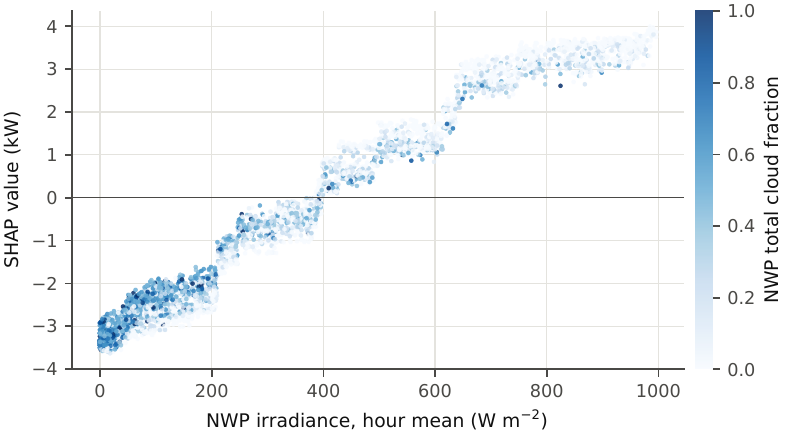}
  \caption{SHAP dependence of the forecast on hour-mean NWP irradiance, coloured by forecast total cloud fraction.}
  \label{fig:shapdep}
\end{figure}

\begin{figure}[htbp]
  \centering
  \includegraphics[width=0.85\textwidth]{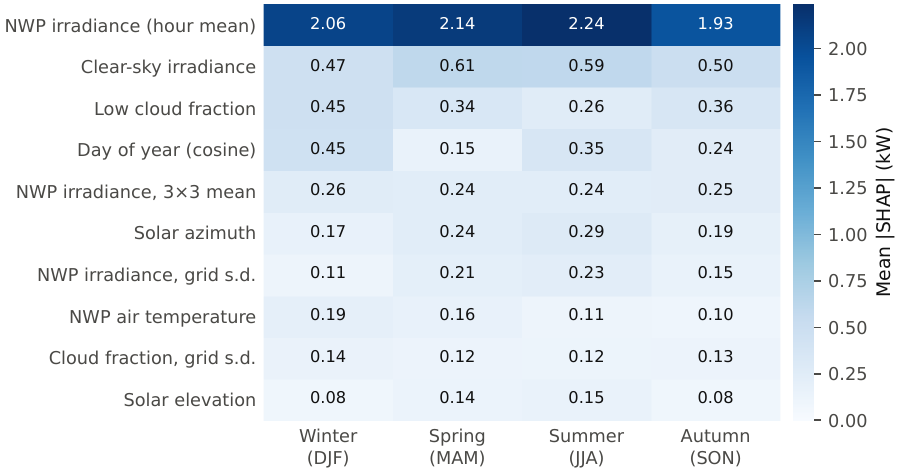}
  \caption{Mean $|\phi_j|$ of the ten most important features by season.}
  \label{fig:shapseason}
\end{figure}

The dependence plot in \cref{fig:shapdep} shows that the effect of irradiance is close to monotonic. At low forecast irradiance, hours with a high forecast cloud fraction receive a less negative contribution, which may mean that the model corrects part of a low bias in NWP irradiance when cloud is forecast. Splitting the SHAP values by season (\cref{fig:shapseason}) shows that low cloud matters most in winter (0.45\,kW against 0.26\,kW in summer), while solar azimuth and the spatial spread of irradiance matter most in summer. Temperature plays a modest part, larger in winter (0.19\,kW) than in autumn (0.10\,kW).

Two checks were run to assess how far these explanations can be trusted. For stability, the model was retained on five expanding windows ending between July 2014 and July 2015 and the SHAP rankings were recomputed on the same test hours. The rankings agreed closely (mean Kendall $\tau=0.89$, minimum 0.83) and the top three features were the same in every window.

For correlated inputs, the four irradiance-related features were examined. The hour-mean irradiance correlates at $r=0.96$ with both the instantaneous value and the $3\times3$ mean, and at $r=0.81$ with clear-sky irradiance. Removing the four redundant features (instantaneous irradiance, the $3\times3$ mean, clear-sky irradiance and the NWP clear-sky index) left accuracy essentially unchanged (11.85\% against 11.79\%). Their credit, however, moved elsewhere: solar elevation, which ranked low in the full model, became the second most important feature (\cref{tab:importance}).

\begin{table}[htbp]
\centering
\caption{Global importance of selected features under three measures.}\label{tab:importance}
\small
\begin{tabularx}{\textwidth}{@{}X >{\centering}p{2.6cm} >{\centering}p{2.6cm} >{\centering\arraybackslash}p{2.6cm}@{}}
\toprule
Feature & Mean $|\phi_j|$, full model (kW) & $I_j$, full model (kW) & Mean $|\phi_j|$, reduced model (kW) \\
\midrule
NWP irradiance (hour mean) & 2.10 & 1.80 & 2.03 \\
Clear-sky irradiance & 0.55 & 0.25 & removed \\
Low cloud fraction & 0.34 & 0.15 & 0.36 \\
Day of year (cosine) & 0.28 & 0.06 & 0.31 \\
Solar elevation & 0.12 & 0.02 & 0.76 \\
Power at the same hour on day $D-1$ & 0.11 & 0.02 & 0.21 \\
\bottomrule
\end{tabularx}
\end{table}

This is the pitfall described in \cref{sec:pitfalls}. The model needs to know how much sunshine is possible at a given hour, and it can learn that equally well from clear-sky irradiance or from solar elevation. SHAP reports whichever one the model happened to use. A statement such as ``clear-sky irradiance is the second most important driver of PV output'' is therefore a claim about one fitted model, not about the plant.

\subsection{Model footprint}\label{sec:footprint}

Forecasts are increasingly run close to the plant, on gateways, inverters or embedded controllers. \Cref{tab:footprint} compares model size and the time taken to predict one sample, measured on a two-core cloud CPU. The figures are meant for comparison between models, not as targets for any particular device.

\begin{table}[htbp]
\centering
\caption{Model size, single-sample prediction time and accuracy.}\label{tab:footprint}
\small
\begin{tabular}{@{}l S[table-format=2.3] S[table-format=2.1] S[table-format=2.2]@{}}
\toprule
Model & {Size (MB)} & {Prediction time (ms)} & {nRMSE (\%)} \\
\midrule
Linear regression & 0.001 & 0.7 & 12.32 \\
Decision tree & 0.02 & 0.7 & 12.99 \\
MLP & 0.28 & 1.0 & 11.61 \\
LightGBM & 0.60 & 0.7 & 11.79 \\
XGBoost & 0.69 & 1.8 & 11.66 \\
SVR & 1.40 & 1.5 & 11.80 \\
Random Forest & 17.7 & 54 & 11.71 \\
Extra Trees & 37.2 & 56 & 11.62 \\
\bottomrule
\end{tabular}
\end{table}

Extra Trees was about 60 times larger than LightGBM and about 80 times slower per prediction, for an accuracy difference of 0.17 points. On memory-limited hardware such as microcontrollers or field-programmable gate array (FPGA) soft cores, a compact boosted model or a small MLP is clearly the better choice.

\subsection{Lessons for practitioners}

Several practical lessons emerge from this case study.
\begin{enumerate}
\item Feature design and time alignment deserve attention before model choice. Physically informed features improved nRMSE by about 2.8 percentage points, whereas switching among the best models changed it by only about 0.2.
\item Validation should follow time order, with a full year held out for testing. Shuffled validation gave an optimistic result here, though only slightly; the gap is likely to be larger when inputs carry more shared information between neighbouring hours.
\item Skill should always be reported against smart persistence and against a simple regression on NWP irradiance. Without these references it is impossible to judge how much the machine learning model actually contributes.
\item AutoML is best used as a search tool. Its final choice depended on the time budget, and a tuned model zoo performed better.
\item Probabilistic forecasts need calibration. The raw 80\% quantile intervals covered only 64.5\% of observations; conformal calibration restored coverage to 81.7\%, at the cost of a moderate increase in interval width (from 3.45 to 4.07,kW).
\item Explanations should be checked before they are acted on. SHAP rankings remained stable over time but changed once correlated inputs were removed, and the largest errors occurred on overcast hours that the NWP failed to anticipate.
\end{enumerate}
\section{Deployment challenges and open problems}\label{sec:challenges}

Moving a forecasting model from a study into daily use raises problems that accuracy tables do not reveal. Six of them deserve attention.

\paragraph{Small data and new sites.} A new plant has no history. Transfer learning from nearby plants, models trained on many sites at once \citep{simeunovic2022}, physical model chains \citep{mayer2021} and time-series foundation models \citep{ansari2024,das2024} all offer a way to start, but little is known about how well they transfer between climates. Reporting results on public benchmark datasets, as done in \cref{sec:case}, would make such comparisons possible.

\paragraph{Non-stationarity.} Plants degrade, panels get dirty, inverters are replaced, and NWP providers update their models. Each of these changes alters the relationship the model has learned. Drift detection and scheduled retraining \citep{gama2014}, with rolling-origin evaluation as the standard test, belong in every deployment. Climate change adds a slower shift. The weather a plant meets over a 25-year life may fall outside its training data, which matters most for tree models because they cannot extrapolate.

\paragraph{Extreme and rare events.} Heat waves, dust and smoke episodes and snow cover produce the largest errors and often the highest costs. They are rare in training data and are smoothed away by average error measures. Evaluation focused on such events, and probabilistic forecasts that widen in unusual conditions, deserve more attention than they receive.

\paragraph{Edge deployment.} Running forecasts on gateways, inverters or embedded controllers cuts latency and communication and keeps data on site. \Cref{sec:footprint} showed that compact boosted trees and small neural networks match the accuracy of large forests at a small fraction of their size. Embedded ML frameworks such as TensorFlow Lite Micro \citep{david2021}, benchmark suites such as MLPerf Tiny \citep{banbury2021}, and implementations on FPGA and system-on-chip platforms make this practical. Designing the model and the hardware together for energy forecasting remains largely unexplored.

\paragraph{Data governance and privacy.} Plant-level data are commercially sensitive, and household PV and consumption data are personal. Federated learning and privacy-preserving analytics allow models to learn across sites without pooling raw data. The chapter on cybersecurity, privacy and trustworthy AI in this volume treats these questions in more depth.

\paragraph{Resource assessment.} For long-term assessment the main AI tasks are correcting satellite and reanalysis estimates against ground measurements, filling gaps, and turning climate projections into site-level yield scenarios \citep{sengupta2021}. Here explainability serves a different audience: lenders and engineers who need to understand the uncertainty behind P50 and P90 estimates. How to connect explainable models with bankable uncertainty estimates is still an open question.

\section{Conclusion}\label{sec:conclusion}

This chapter reviewed AI methods for PV power forecasting and resource assessment and set out a reproducible workflow that combines automated model search with explainable AI. On an open day-ahead dataset from Porto, the best ML models halved the error of smart persistence and reached an nRMSE of about 11.6\% of installed capacity over a full test year. The case study also showed where effort pays off: in physically informed features and careful time alignment rather than in the choice among strong algorithms, in validation that respects time and in reference baselines, in calibrated uncertainty, and in explanations that are tested for stability and for the effect of correlated inputs.

The presented results also refine some conclusions from the earlier study on Malaysian data \citep{bakht2025}. Extra Trees was again among the most accurate models, but well-tuned neural networks and boosted trees performed equally well, and a high SHAP ranking for a single input can reflect the fitted model as much as the plant itself. In future work, the workflow should be tested across a wider range of climates and plant types and compared with physical model chains and foundation models on shared benchmarks. Developing compact, explainable models that can run on edge hardware is another priority.

\section*{Acknowledgements}
The authors are grateful to INESC TEC for making the PV measurements and NWP data publicly available \citep{andrade2020data}, and to MeteoGalicia for providing the numerical weather predictions. 

\section*{Data and code availability}
The dataset can be obtained from the INESC TEC Research Data Repository (\url{https://doi.org/10.25747/edf8-m258}). All analyses were performed in Python using scikit-learn \citep{pedregosa2011}, XGBoost, LightGBM, FLAML, SHAP and pvlib, with a fixed random seed. The code and complete results are available from the corresponding author on request. 
\section*{Declaration of generative AI use}
The authors used Claude (Anthropic) for language editing and checking references. However, all the content was reviewed, revised and verified by the authors, who take full responsibility for the chapter.

\bibliographystyle{unsrtnat}
\bibliography{references}

\end{document}